\documentclass[conference]{IEEEtran}
\IEEEoverridecommandlockouts
\usepackage{cite}
\usepackage{amsmath,amssymb,amsfonts}
\usepackage{algorithmic}
\usepackage{graphicx}
\usepackage{textcomp}
\usepackage{xcolor}
\usepackage{framed}
\def\BibTeX{{\rm B\kern-.05em{\sc i\kern-.025em b}\kern-.08em
    T\kern-.1667em\lower.7ex\hbox{E}\kern-.125emX}}
\begin{document}

\bstctlcite{IEEEexample:BSTcontrol}

\title{Self-Learning Tuning for Post-Silicon Validation\\

\thanks{This research was supported by Advantest as part of the Graduate School "Intelligent Methods for Test and Reliability" (GS-IMTR) at the University of Stuttgart.}
}

\author{
	\IEEEauthorblockN{Peter Domanski, Dirk Pflüger}
	\IEEEauthorblockA{\textit{Institute for Parallel and Distributed Systems} \\
		\textit{University of Stuttgart}\\
		Stuttgart, Germany \\
		\{peter.domanski, dirk.pflueger\}@ipvs.uni-stuttgart.de} 
	\and
	\IEEEauthorblockN{Jochen Rivoir, Raphaël Latty}
	\IEEEauthorblockA{\textit{Applied Research and Venture Team}\\
		\textit{Advantest Europe GmbH} \\
		Böblingen, Germany \\
		\{jochen.rivoir, raphael.latty\}@advantest.com}
}

\maketitle

\begin{abstract}
Increasing complexity of modern integrated circuits makes design validation more difficult. Existing approaches are not able anymore to cope with the complexity of tasks such as robust performance tuning in post-silicon validation. Therefore, we propose a novel learn-to-optimize approach based on reinforcement learning in order to solve complex and mixed-type tuning tasks in an efficient and robust way.
\end{abstract}


\section{Introduction}
Studies suggest that validation is a major bottleneck in semiconductor integrated circuit (IC) design that occupies up to 70\% of the time, efforts, and resources that are used during the design process \cite{mishra2017post}. In general, design validation can be divided into three main stages \cite{mishra2019post}: pre-silicon validation, post-silicon validation (PSV), and in-field debugging. 
In the following, we focus on PSV. More specifically, we describe the task of robust performance tuning and our motivation for 
a novel approach to this task. \newline 

Today, PSV is largely considered as an art offering only few systematic solutions. Moreover, the rising complexity of modern ICs increases the difficulty of PSV even further. Existing approaches are not able to cope with the complexity of future systems, especially given a limited time budget. 
Therefore, PSV is an exciting research field offering novel opportunities and challenges \cite{mitra2010post}. Ensuring functional correctness, checking security properties, and meeting performance constraints are important tasks in PSV. Besides, robustness, e.g., against process variations, is crucial. 
In order to ensure the best possible performance and robustness, designs rely on tuning to compensate impacts of process variations and of non-ideal design implementations. Tuning consists in setting variables and registers, so-called "tuning knobs" such as bias settings or adjustable currents and voltages, which may be adjusted as a function of given (operating) conditions, e.g., temperature or operating mode of an IC. In addition, the number of tuning knobs on ICs is rising because affordability is increasing in modern technologies. \newline 

Setting tuning knobs such that device parameters stay within specification limits and optimize performance goals still remains a complicated, bound-constrained, mixed-type optimization task. 
Depending on the different data types of the variables, a first naive approach is to use either gradient methods or more flexible derivative-free optimization strategies. These strategies are important in PSV because there is no analytical expression describing the true behavior of manufactured ICs (similar to black-box functions). Thus, we have no access to higher-order moments of the underlying function. 
Applying state-of-the-art optimization strategies to solve 
mixed-type 
optimization problems results in approximate, point-wise solutions because exact solutions are intractable. Given the tight schedule in PSV, point-wise solutions remain very inefficient. To avoid these pitfalls, we propose a new self-learning tuning approach that aims to learn a mapping from conditions to tuning knobs, which we call tuning law. 
Such a tuning law is flexible, robust, and efficient as it can be applied either in IC-specific or IC-independent setups. In the following sections, we show how our novel approach relates to existing work and describe how to learn a 
tuning law with a runtime fast enough even in high-dimensional setups and with the increasing complexity of modern ICs. Finally, we show preliminary results of our approach and discuss the experiments and future work. 


\section{Related work}
\textbf{Learn-to-optimize} \quad Learn-to-optimize is a data-driven approach to learn a model that aims to replace hand-crafted optimization algorithms (e.g., SGD, RMSprop, and Adam). Many existing works, e.g., \cite{andrychowicz2016learning, chen2017learning, lv2017learning, wichrowska2017learned, zoph2016neural, li2016learning} leverage Recurrent Neural Networks (RNN) as (coordinate-wise) optimizers and rely on gradient information of the objective function to output parameter updates. Architectural details of the learned optimizer model (e.g., Long Short Term Memory (LSTM) networks, and hierarchical RNN \cite{wichrowska2017learned}), training methods (e.g., Truncated Backpropagation Through Time, reinforcement learning \cite{chen2017learning, zoph2016neural}, and meta-learning), and input features can differ, but the main focus is on continuous optimization problems, especially on training of Deep Neural Networks. Besides learning the full update rule, learn-to-optimize was customized to automatic hyperparameter tuning (auto-tuning), neural architecture search, and adversarial attack scenarios.
\newline

\textbf{Hyperparameter tuning} \quad A large number of hyperparameter optimization methods have been proposed to improve the performance of learning algorithms. Many frameworks, e.g., \cite{akiba2019optuna, liaw2018tune, bergstra2013hyperopt} exist that support most common methods like Grid Search, Random Search, Bayesian optimization, Gradient-based optimization, as well as Evolutionary- and Population-based optimization. These methods differ in efficiency and assumptions with respect to properties of the objective function. In PSV, methods like Grid- or Random Search are too inefficient to be used in practice. Likewise, Bayesian- and Gradient-based optimization are difficult to use due to assumptions like smoothness or differentiability of the objective function. Such assumptions do not necessarily hold in PSV, neither are properties of the objective function known beforehand.

\section{Methodology}
\begin{figure*}
	\centering
	\includegraphics[width=0.95\textwidth]{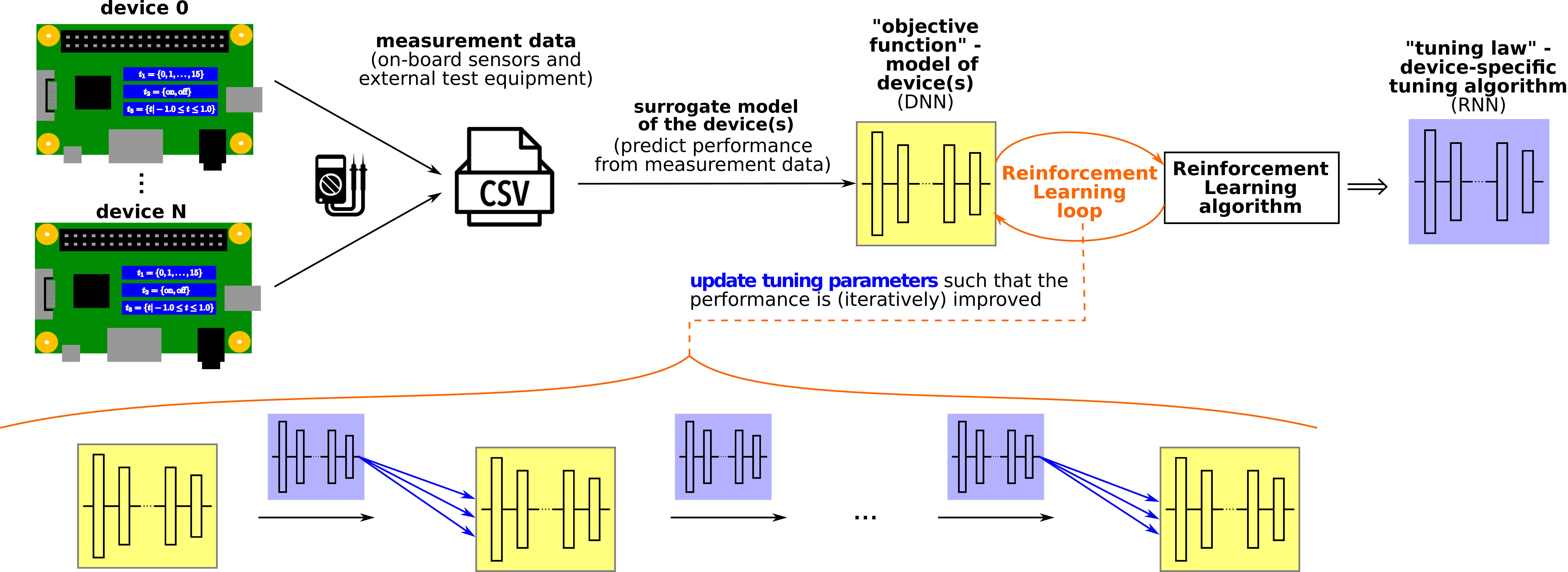}
	\caption{Proposed approach to self-learning tuning in PSV. The tuning law for the surrogate objective function (yellow) is learned iteratively within the reinforcement learning loop (orange). In this application, we use a performance measure (aggregated from measurement outputs) of the device to train and evaluate a (device-specific) tuning algorithm in form of a (Recurrent) Neural Network (blue) on the surrogate model. After training, the tuning algorithm runs on real-world devices. Note that the complexity of surrogate models for the objective function could be heavily reduced by using variable selection methods (e.g., \cite{yiwen_imtr} a recent work from the GS-IMTR) due to high dimensional input data in PSV.}
	\label{methodology}
\end{figure*}

In order to learn a tuning law, we are following the path of learn-to-optimize and leverage reinforcement learning methods as in \cite{chen2017learning, zoph2016neural} to train an optimization algorithm. In contrast to hand-crafted methods, this data-driven approach automatizes the design of efficient optimization methods by observing their own execution performance. Whereas most investigations focus on continuous optimization, we make use of reinforcement learning in the training procedure to enable learning of optimization methods for mixed-type problems. Similar to \cite{chen2017learning}, the objective function is a black box. In our application, the objective is solely described by the given data set. Instead of optimizing network parameters, we aim to find inputs that produce the best outputs (in terms of device performance) without relying on gradients or approximations of gradients. This resembles the scenario of black-box adversarial attacks, e.g. in \cite{guo2019simple}. Fig.~\ref{methodology} shows an overview of the proposed approach. As illustrated, we iteratively train an optimization algorithm on a surrogate model in the outer reinforcement learning loop. The learned optimizer itself (iteratively) improves the performance measure of the surrogate in the inner loop. The performance improvement is used as feedback to the outer training loop. The surrogate model allows us to formulate tuning as an iterative optimization process and train the tuning law efficiently within the reinforcement learning loop. \newline

\begin{figure}
	\centering
	\hspace*{0.5cm}
	\includegraphics[width=0.5\textwidth]{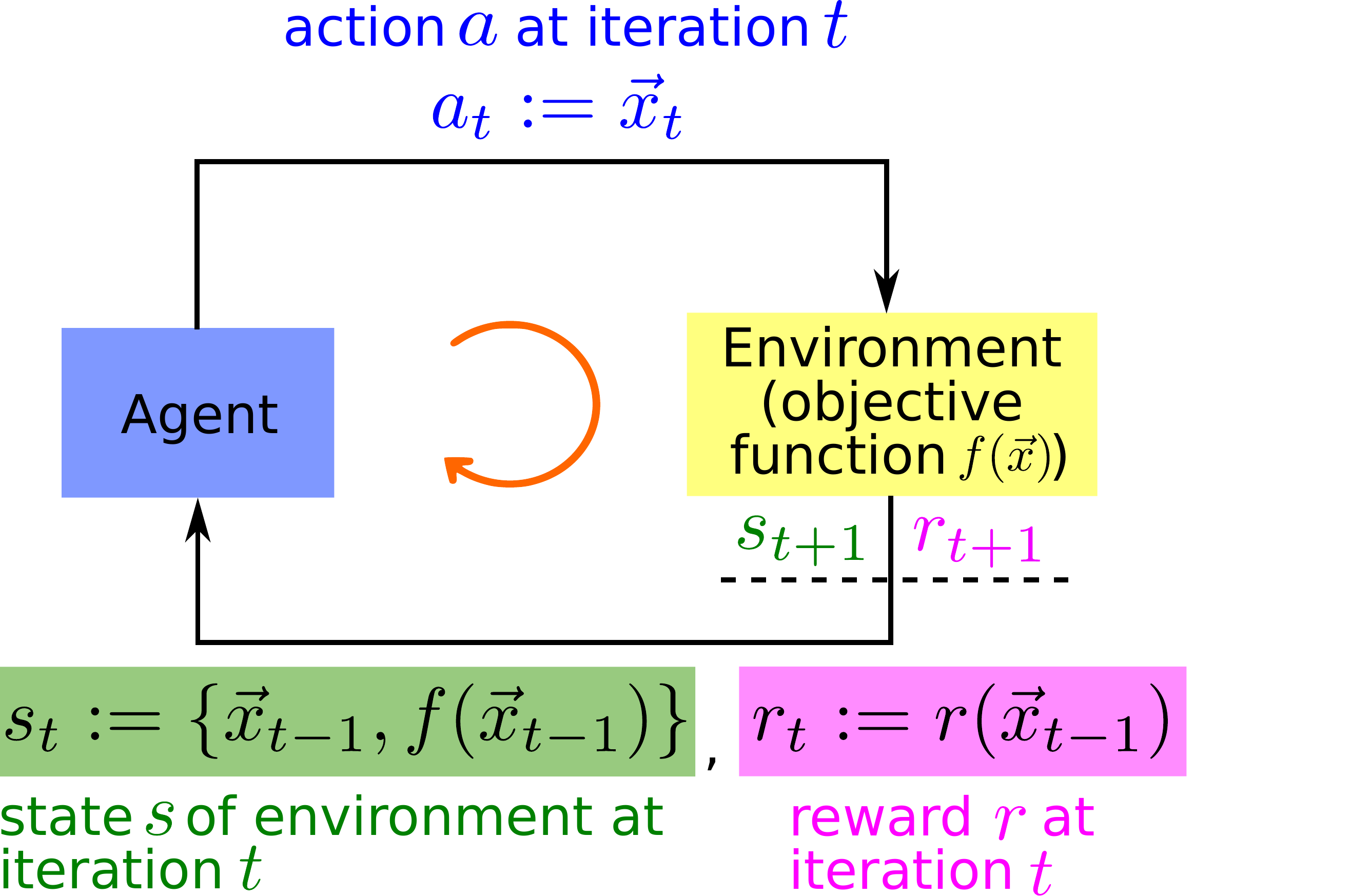}
	\caption{Learn-to-optimize approach. The visualization shows key components (agent, environment, state- and reward definitions) in the reinforcement learning loop (orange).}
	\label{rl_opt_loop}
\end{figure}

The general reinforcement learning setting of learn-to-optimize is shown in Fig.~\ref{rl_opt_loop}. In this setting, an agent chooses an action, e.g. optimization parameter values $\vec{x}_{t}$ at each iteration $t$, which changes the environment, e.g., the objective function value $f(\vec{x}_{t})$. The agent receives feedback, typically in the form of a reward, based on the consequences of the action. The goal of the agent is to choose a sequence of actions based on observations of the environment's state in a way that the agent maximizes the cumulative reward over all iterations \cite{li2016learning}. The final goal is to find $\vec{x}^*=\arg\max f(\vec{x})$. The reward definition we use reflects these optimization goals.  
The computations of our methodology can be summarized by the following steps: 

\begin{framed}
\begin{itemize}
	\item[1.] Given current state $\vec{s}_{t,\text{NN}}:=\{(\vec{x}_{t-1},f(\vec{x}_{t-1}))\}$ or \newline $\vec{s}_{t,\text{RNN}}:=\{(\vec{x}_{t-1}, f(\vec{x}_{t-1}), \vec{h}_{t-1})\}$ \newline propose update $\vec{x}_{t}$  \newline
	($\vec{h}_{t-1}$: internal state of RNN at time step $t-1$)
	\item[2.] Observe response of environment, e.g., $f(\vec{x}_{t})$
	\item[3.] Update internal statistics to produce $\vec{s}_{t+1}$ and $r_{t+1}$
\end{itemize}
\end{framed}

The update rule (e.g., agent) in 1. is defined by a (Recurrent) Neural Network parameterized by $\theta$, such as $f_{\theta,\text{(R)NN}}:= \vec{s}_t \rightarrow \vec{x}_{t}$. 
The architecture consists of a two-layer LSTM network with tanh activation units. The parameters $\theta$ are updated by applying reinforcement learning algorithms, e.g., REINFORCE to maximize cumulative reward. \newline

Given a similar computing budget, our approach can have superior performance showing much faster convergence speed compared to classical methods, in particular for difficult (e.g. non-convex) optimization problems. Thus, learning to optimize has the potential to overcome the limits of the actual analytical methods in PSV. Another benefit is that the solutions of reinforcement learning can be directly used as a tuning law. \newline


\section{Experiments}
In this section, we provide preliminary results of our methodology. In particular, we use a real-world data set provided by Advantest that consists of data from multiple devices. The objective function $f$  we aim to optimize is an aggregation of device-specific models, e.g., Neural Networks that describe the input-output behavior of each device. The aggregation is designed to cover the behavior across devices. The parameter $\vec{x}$ we optimize are integer values, namely the tuning knobs of the devices. The goal is to find the best value $\vec{x}^*$ which corresponds to the best possible configuration of the tuning knobs. In this preliminary experimental setup, we have not specified any tuning conditions. Thus, the tuning law we aim to learn is a constant function. However, our methodology can be extended to support arbitrary data types and condition-dependent tuning laws. \newline

\begin{figure}
	\centering
	\includegraphics[width=0.5\textwidth]{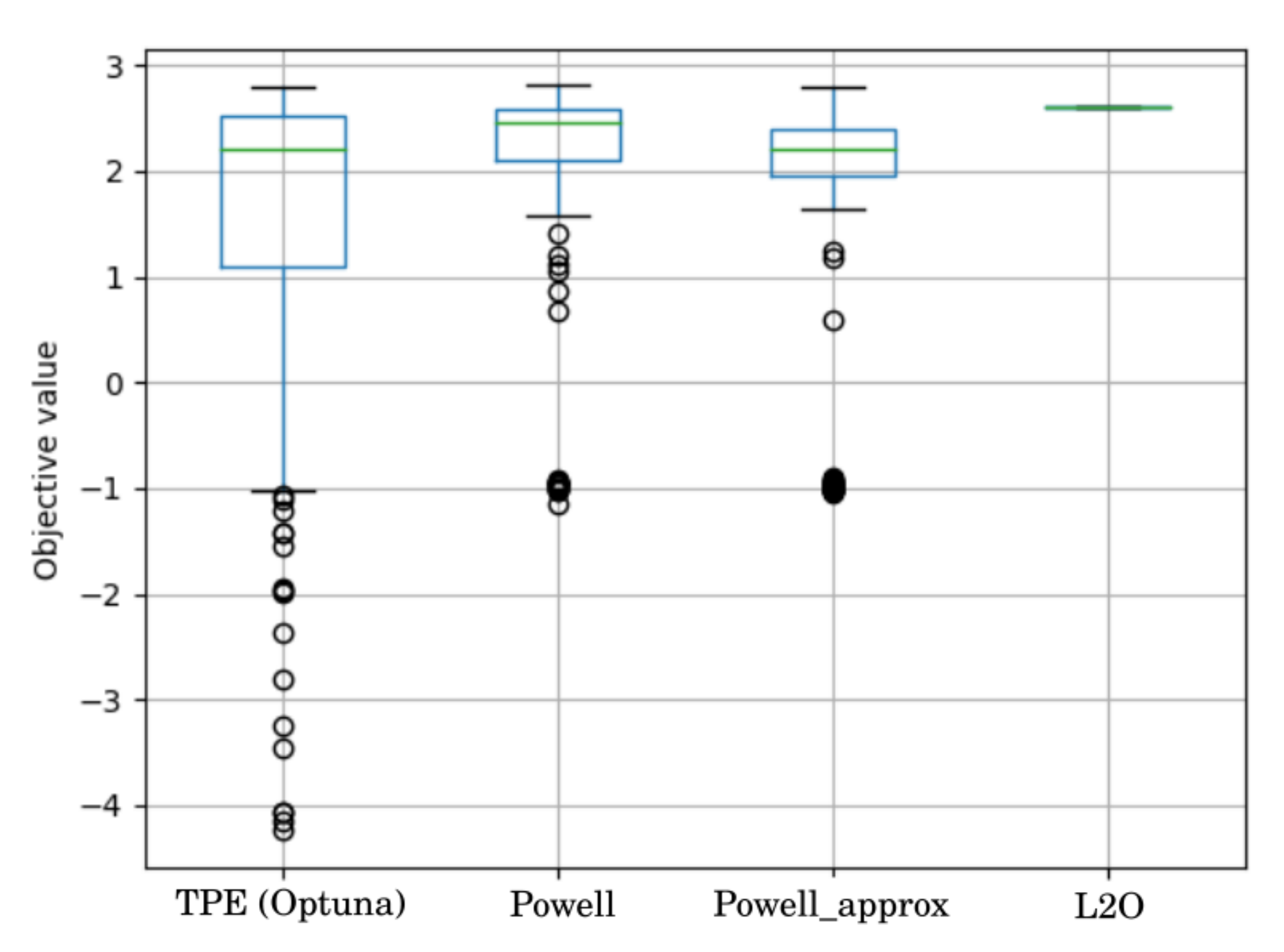}
	\caption{Box-plot visualization of the distributions of achievable objective function values. We compare Powell's method with default hyperparameters, Powell's method with a comparable computational budget, TPE (Optuna), and learn-to-optimize (our method). For each evaluation of the objective function values, we use 16 random initial parameter values.}
	\label{results_all_methods}
\end{figure}

Fig.~\ref{results_all_methods} shows the objective function value $f(\vec{x}^*)$ of the described experimental setup after training. We compare our approach (L2O) with a state-of-the-art optimization algorithm called Powell's method. In previous experiments, this method has proven to be the best in our experiments compared to other applicable (hyperparameter tuning) algorithms, e.g., Evolutionary- and Population-based approaches. Here, we compare two versions of Powell's method, a version with default hyperparameters and another with a comparable computational budget to our approach. As a baseline, we use the Tree-structured Parzen Estimator (TPE) algorithm \cite{bergstra2011algorithms} with a median pruner (early stopping of unpromising trials) as implemented in Optuna \cite{akiba2019optuna}. Similar to both versions of Powell's method, we set the number of maximum trials for TPE such that computational budgets are comparable. \newline 

\begin{figure}
	\centering
	\includegraphics[width=0.5\textwidth]{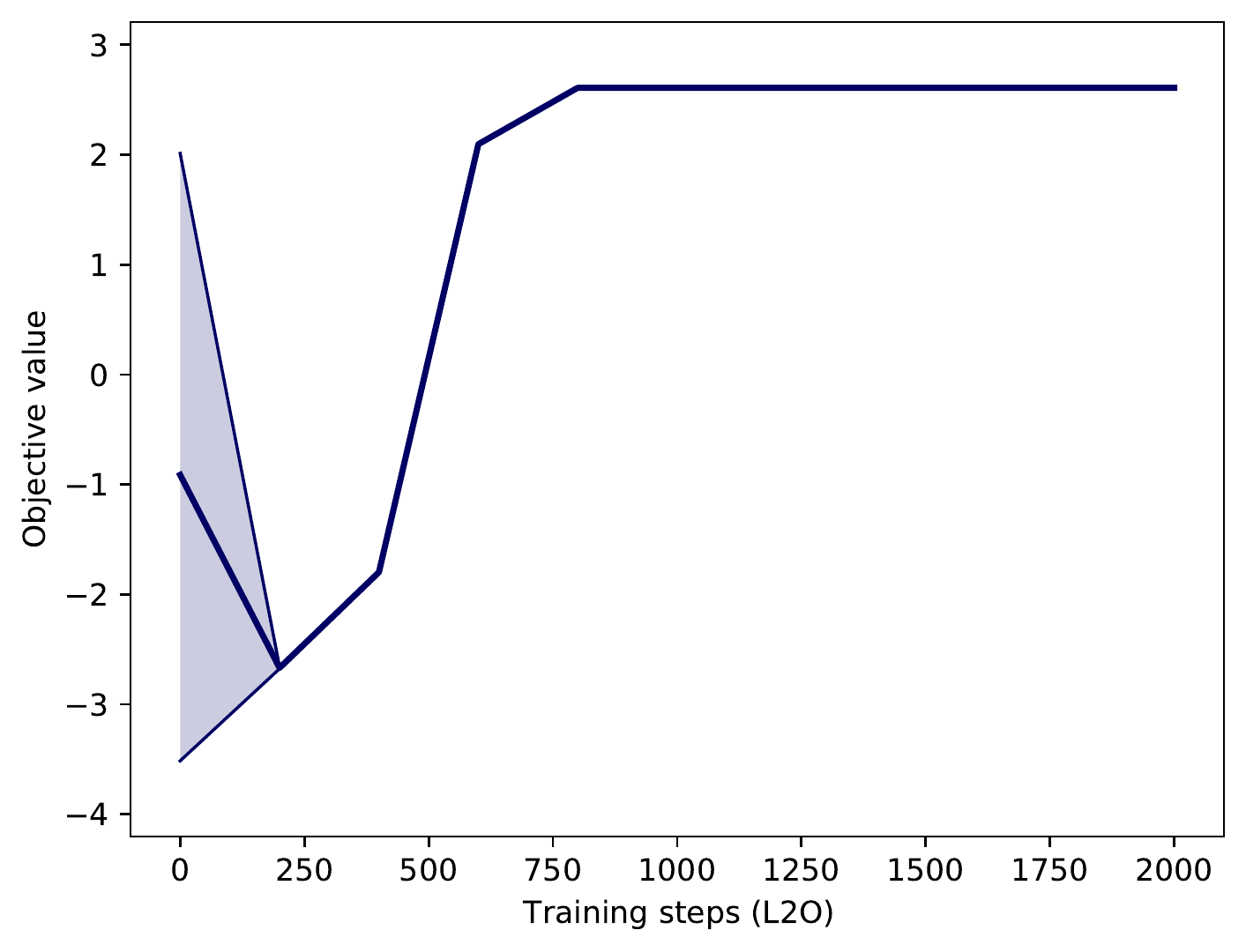}
	\caption{Performance (objective value) of the agent (RNN that we use as tuning law) over the reinforcement learning process. Training the agent for 2000 steps with the learn-to-optimize approach took 1.5 hours on our hardware. For each evaluation, we use 16 random initial parameter values.}
	\label{results_L2O}
\end{figure}

For both versions of Powell's method, we observe a constant variability in the objective function value caused by different initial values. Moreover, the average performance does not improve with time. The TPE baseline looks similar but shows a much larger variability in the objective function value. In contrast, our approach benefits from additional effort during training and improves over time, see Fig.~\ref{results_L2O}. Thus, our methodology results in a very stable tuning law almost independent of initial parameter values and reaches a high performance. Tab.~\ref{speed_up} shows the improvements in computation time. 

\section{Conclusion}

To summarize, the learn-to-optimize approach has appealing properties and opportunities that match the characteristics of the tuning task in PSV very well, e.g.,  the high-dimensional mixed-type problems. First solutions outperform state-of-the-art, point-wise optimization results in final performance, efficiency, and convergence speed. Finally, our approach allows us to learn the tuning on actual devices by taking advantage of the data related to the objective function of interest. There is no need to introduce new a priori assumptions regarding the objective nor additional hyperparameters of the resulting optimization algorithm. In the future, we aim to study more complicated objective functions (e.g., min-max) and to learn tuning laws that depend on given conditions. The latter means that we aim to optimize tuning parameters as a function of conditions, such as currents, voltages or, the operating mode of devices. With state-of-the-art optimization methods, this gets infeasible as we have to re-run the entire optimization procedure for all possible values of the conditions, which results in very high time and memory consumption.

\begin{table}
	\begin{center}
		\resizebox{0.4\textwidth}{!}{\begin{tabular}{|c|c|}
				\hline
				 &  Computation time (avg.)\\
				\hline
				 L2O & $\mathbf{1.118 \, \textbf{s}}$ \\
				 \hline
				 TPE (Optuna) & $108.237 \, \text{s}$ \\
				 \hline
				 Powell (default) & $129.684 \, \text{s}$ \\ 
				 \hline
				 Powell (approx.) & $39.196 \, \text{s}$ \\
				\hline
		\end{tabular}}
		\vspace*{0.25cm}
		\caption{Time to optimize (without training time).}
		\label{speed_up}	
	\end{center}	
\end{table}

\newpage

\bibliographystyle{IEEEtran}
\bibliography{IEEEabrv,tuz_literature}

\end{document}